\documentclass[review]{elsarticle}
\graphicspath{ {figure_pdf/} }
\usepackage[hyphens]{url}
\usepackage{hyperref}
\usepackage{float}
\usepackage{verbatim} %comments
\usepackage{apalike}
\usepackage{amsmath}
\restylefloat{figure}
\restylefloat{table}

\journal{Expert Systems with Applications}

\biboptions{authoryear}

\begin{document}
\begin{frontmatter}

% \begin{titlepage}
% \begin{center}
% \vspace*{1cm}

% \textbf{ \large LA-HCN: Label-based Attention for Hierarchical Multi-label Text Classification Neural Network}

% \vspace{1.5cm}

% % Author names and affiliations
% Xinyi Zhang$^a$ (xinyi001@e.ntu.edu.sg), Jiahao Xu$^a$ (jiahao004@e.ntu.edu.sg), Charlie Soh$^a$ (csoh004@e.ntu.edu.sg), Lihui Chen$^a$ (elhchen@ntu.edu.sg) \\

% \hspace{10pt}

% \begin{flushleft}
% \small  
% $^a$ Nanyang Technological University, Singapore

% \begin{comment}
% Clearly indicate who will handle correspondence at all stages of refereeing and publication, also post-publication. Ensure that phone numbers (with country and area code) are provided in addition to the e-mail address and the complete postal address. Contact details must be kept up to date by the corresponding author.
% \end{comment}

% \vspace{1cm}
% \textbf{Corresponding Author:} \\
% Assoc Prof Chen, Lihui (Dr.) \\
% Center for Info Sciences and Systems \\
% School of Electrical and Electronic Engineering \\
% Nanyang Technological University \\
% 50 Nanyang Avenue, Singapore 639798 \\
% Republic of Singapore \\
% Tel: +65 6790 4484 \\
% Email: elhchen@ntu.edu.sg

% \end{flushleft}        
% \end{center}
% \end{titlepage}

\title{LA-HCN: Label-based Attention for Hierarchical Multi-label Text Classification Neural Network}

\author[label1]{Xinyi Zhang}
\ead{xinyi001@e.ntu.edu.sg}

\author[label1]{Jiahao Xu}
\ead{jiahao004@e.ntu.edu.sg}

\author[label1]{Charlie Soh}
\ead{charliesoh@ntu.edu.sg}

\author[label1]{Lihui Chen}
\ead{elhchen@ntu.edu.sg}

\cortext[cor1]{Corresponding author.}
\address[label1]{Nanyang Technological University, Singapore}

\begin{abstract}
Hierarchical multi-label text classification (HMTC) has been gaining popularity in recent years thanks to its applicability to a plethora of real-world applications. The existing HMTC algorithms largely focus on the design of classifiers, such as the local, global, or a combination of them. However, very few studies have focused on hierarchical feature extraction and explore the association between the hierarchical labels and the text. In this paper, we propose a \textbf{L}abel-based \textbf{A}ttention for \textbf{H}ierarchical Mutlti-label Text \textbf{C}lassification Neural \textbf{N}etwork  (\textbf{LA-HCN}), where the novel label-based attention module is designed to hierarchically extract important information from the text based on the labels from different hierarchy levels. Besides, hierarchical information is shared across levels while preserving the hierarchical label-based information. Separate local and global document embeddings are obtained and used to facilitate the respective local and global classifications. In our experiments, LA-HCN outperforms other state-of-the-art neural network-based HMTC algorithms on four public HMTC datasets. The ablation study also demonstrates the effectiveness of the proposed label-based attention module as well as the novel local and global embeddings and classifications. By visualizing the learned attention (words), we find that LA-HCN is able to extract meaningful information corresponding to the different labels which provides explainability that may be helpful for the human analyst.
\end{abstract}


\end{frontmatter}

\section{Introduction}
\label{introduction}

\label{sec:introduction}
In recent years, there has been a growing interest in hierarchical multi-label classification (HMC) which can be applied in a wide range of applications such as International Patent Classification \citep{ipcsurvey'2014}, product annotation \citep{CapsNetwork'2019} and advertising recommendation \citep{advertiser'2013}. In the common \textit{flat} classification problem, each input sample is only associated with a single label from a set of disjoint labels. However, in HMC problem, the labels are organized in the form of a tree or a Directed Acyclic Graph(DAG) \cite{hmcsurvey'2011} and each input sample is usually associated with multiple labels, which made it more challenging.

The most straight-forward approach in dealing with HMC problem is to convert it to a flat multi-label classification problem by simply ignoring the relevance between labels \citep{deeppatent'2018,patentkeyword'2018,CapsNetwork'2019}. The main disadvantage in doing so is the loss of the useful hierarchical information. Alternatively, the \textit{local} approach \citep{fewwords'1997} is designed to perform multi-label classification, where the classifications are carried out at each level of the label hierarchy (e.g., Local Classifier per Parent Node, Local Classifier per Level and Local Classifier per Node). The overall classification results are then generated based on these local predictions. While hierarchical information can be better utilized in \textit{local} approaches,  misclassifications are easily propagated to the next levels \citep{punera2008enhanced}. \textit{Global} approaches are proposed to learn a single global model for all labels to reduce the model size and consider the entire label hierarchy at once \citep{functional'2005,kiritchenko2006learning}. These \textit{global} classifiers are typically built on \textit{flat} classifiers with modifications made to integrate the hierarchical information of labels \citep{large'2009,vens2008decision} into the model. Recently, more algorithms which combine the \textit{local} and \textit{global} approaches are proposed \citep{hmcn'2018,rl'2019}.

All algorithms introduced above only focus on the design of hierarchical classifier while ignoring the hierarchical features which may be extracted and they are important in HMC as well. \citet{HARNN'2019} and \citet{rojas2020efficient} consider hierarchical feature extraction in their work. However, the extraction process is designed and fulfilled by applying the typical attention mechanism over the whole text. Since in HMC problem the text may be associated with multiple labels at each hierarchy level, the features extracted from typical attention may be diluted.  We believe it is reasonable to hypothesize that a label-based attention, where information extraction is performed based on different labels at different hierarchical levels, would allow the model to be more interpretable and have an overall better performance in accuracy. Given the above motivations, we propose LA-HCN --- a HMTC model with a label-based attention to facilitate label-based hierarchical feature extraction, where we introduce the concept and mechanism of \textit{component} which is an intermediate representation that helps bridge the latent association between the words and the labels for label-based attention.

\paragraph{Contribution} Main contributions of this work:
\begin{itemize}
    \item We propose a novel HMTC model capable of learning disjoint features for each hierarchical level, while sharing the hierarchical information learned across levels. Besides, both local and global classifiers with respective embeddings are applied to reduce the impact of error-propagation across levels.
    \item We propose a novel mechanism to learn label-based word attention at each level such that the important features of each document can be captured based on individual labels and prove that such mechanism can provide more interpretable results.
    \item We evaluate LA-HCN against both classical and state-of-the-art neural network HMTC baselines on four benchmark datasets and LA-HCN outperforms them on all the datasets. The ablation study shows the effectiveness of the component mechanism and different classifiers applied in LA-HCN and demonstrates that the learned label-based attention is able to give reasonable interpretation of the prediction results.
\end{itemize}

\section{Related Work}
HMTC plays an important role in the field of text classification as it is widely applied in many different applications. Some work on text classification focus on text encoding such as Doc2vec \citep{le2014distributed}, LSTM \citep{lstm'1997} and BERT \citep{devlin2019bert} while others pay more attention to classifier design, both of which are critical for HMTC.

There are two main directions in performing hierarchical classification --- local and global approaches \citep{hmcsurvey'2011}. \citet{fewwords'1997} is the first type of local classifier which is proposed to explore the hierarchy by using local information and build multiple local classifiers around it. Following which, a series of local approaches including LCN, LCPN as well as LCL\footnote{LCN, LCPN and LCL denotes local classifier per node, local classifier per parent node and local classifier per level approach respectively.} based methods are proposed \citep{fagni2007selection,secker2010hierarchical,costa2007comparing}. However, local approaches on their own posses high risk of error-propagation. On the other hand, global approaches are designed to build a single classifier to explore hierarchical information globally, thus reducing the overall model size. Most global approaches are generally modified from the flat classification algorithms such as hAnt-Miner \citep{inproceedingsoteros}, \citet{vens2008decision}(decision tree based) and GMNBwU \citep{5360345}(Naive Bayes classifier based). Recently, more neural network based HMC algorithms which combine both local and global approaches are proposed \citep{rl'2019,hmcn'2018}. However, these algorithms ignored the hierarchical feature extraction which is also important for HMC.
 
The most relevant studies on hierarchical feature extraction include HAN \citep{hierattention'2016}, SMASH \citep{smash'2019} and HARNN \citep{HARNN'2019}. However, HAN proposed to extract hierarchical information based on the document structures (i.e., from words to sentences) instead of label structures. SMASH is similar to HAN except they use BERT as the base model. HARNN is the study that is most similar to our work. It extracts hierarchical information at different label levels and generate document embeddings for local and global classifications. However, HARNN applies a general attention to extract document features which may not be as explicit and effective compared with the proposed label-based attention. Besides, the same document embedding is used in both local and global classifications which may affect the performance due to error propagation.

More specifically, the main strengths of LA-HCN over other HMTC algorithms such as HARNN are three folds. Firstly, the LA-HCN employs label-based attention, where the attention is learned based on the individual label. This is in contrary to the typical attention as employed in HARNN, where the attention is shared among labels. Secondly, the information sharing across hierarchy levels is more flexible and effective in LA-HCN.  In HARNN, the features learned at the level $h$ is highly dependent on the features learned in the level $h-1$ (i.e., the features which are not important for the classification at the level $h-1$ will not be considered at the level $h$). On the other hand, the feature learning in LA-HCN is based on the labels at $h$ with the considerations of  the features from $h-1$. In contrast with HARNN, LA-HCN considers even the non-important features in $h-1$ while learning the features for level $h$. Lastly, the design of global and local classification are different in HARNN and LA-HCN. In HARNN, both global and local classification are performed with the same document embedding. While in LA-HCN, the respective embeddings are obtained for global and local classification for better error propagation control.

In our experiments, we demonstrate that the label-based attention can provide more interpretable results and achieve higher classification accuracy.

\section{LA-HCN Framework and Module Design Details}
    
    \paragraph{\textbf{Problem Definition}} Given a set of documents $\mathcal{X} = \{X_1, X_2, \dots, X_K\}$ as well as their corresponding labels $\mathcal{Y} = \{Y_1, Y_2,\dots,Y_K\}$ where each document is represented with a sequence of words $X_i = \{w_1, w_2,\dots,w_N\}$ and $Y_i = \{l_1,l_2,\dots\}$ contains all hierarchical labels associated with the document $X_i$. The objective of this work is to learn a mapping $\mathcal{P}:\mathcal{X} \to \mathcal{L}$ to predict the corresponding labels for each document by analyzing the document content as well as the hierarchical structure information of labels.

    Here, the label hierarchy is represented as $(\mathcal{L}, \leq_{h})$, where $\mathcal{L}$ denotes the set of labels and $\leq_h$ denotes a partial order (which is a tree structure in this work) representing the superclass relationship: $\forall  l_1,l_2 \in \mathcal{L}:l_1\leq_h l_2$ if and only if $l_1$ is a superclass of $l_2$. In the hierarchical classification problem, $\l_i \in Y_i \Rightarrow \forall l_j \leq_h l_i:l_j \in Y_i$. Besides, we also define the level of a label as the number of its direct or indirect superclasses (we set a virtual label \textit{root} as the root of all label hierarchy) and the level of a label $l_i$ is represented as $\psi(l_i) = h$.
    
\begin{figure*}
        \centering
        \includegraphics[width=4.75in]{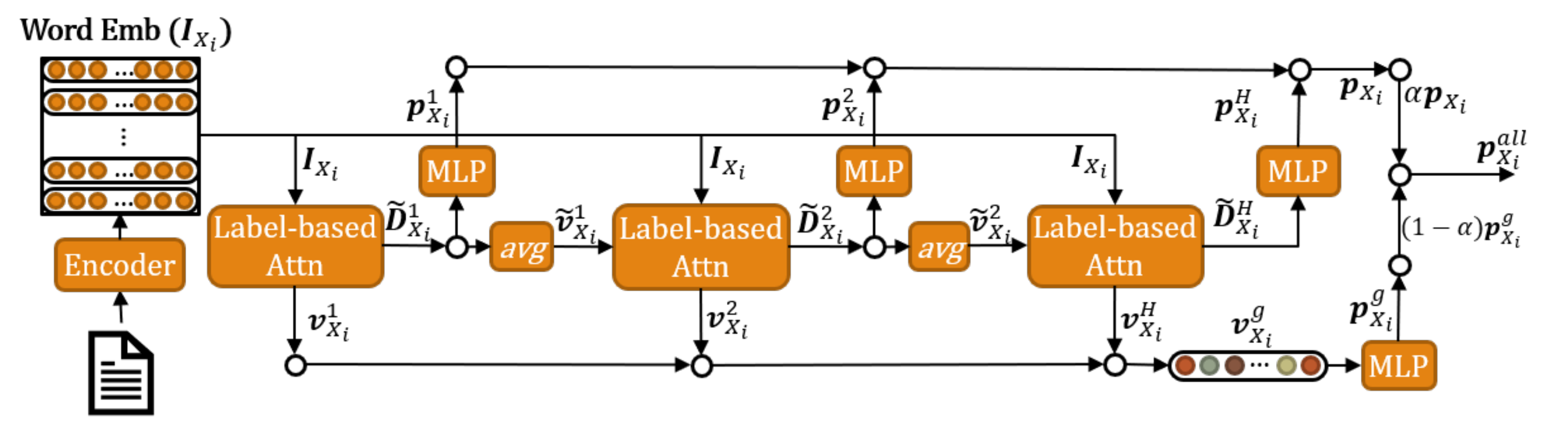}
        \caption{The overall structure of the proposed method. Given the input document $X_i$, a text encoder is applied to get word embeddings ($\boldsymbol{I}_{X_i}$) which are used at all hierarchical levels. At the level $h$, a Label-based Attention Module is applied to generate the processed label-based document embeddings ($\tilde{\boldsymbol{D}}_{X_i}^h$) and the global document embedding ($\boldsymbol{v}^h_{X_i}$) based on $\boldsymbol{I}_{X_i}$ together with the local document embedding ($\tilde{\boldsymbol{v}}_{X_i}^{h-1}$) from the previous level. Then $\tilde{\boldsymbol{D}}_{X_i}^h$ is employed to perform local classification. Global document embeddings ($\boldsymbol{v}^h_{X_i}$) extracted from each level are concatenated to perform global classification. Finally, the prediction results from local and global classifiers are combined to obtain the final results.}  
        \label{fig:overall_structure}
    \end{figure*}

    \paragraph{\textbf{Framework}} The overall structure of LA-HCN is as shown in Fig. \ref{fig:overall_structure}. For each hierarchical level, a label-based attention module is applied to generate both local and global document embeddings for local and global classification, respectively. The details of the 1) Label-based attention module as well as the 2) Classification module will be introduced in the following sections.
    \subsection{Label-based Attention Module}
\begin{figure}
                \centering
                \includegraphics[width=2.5in]{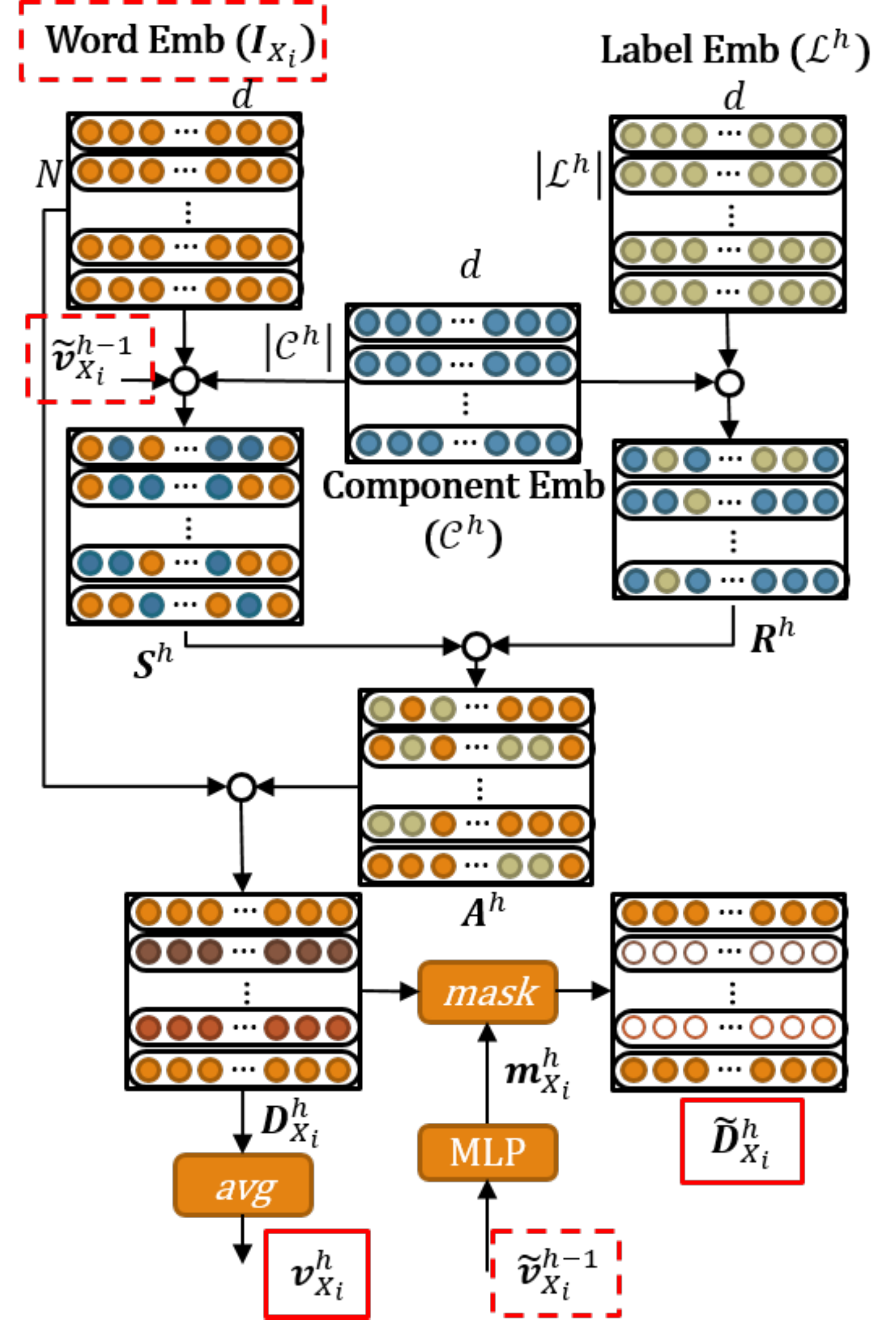}
                \caption{The details of Label-based Attention Module. At the level $h$, a set of component embeddings ($\mathcal{C}^h$) together with word embeddings ($\boldsymbol{I}_{X_i}$) and label embeddings ($\mathcal{L}^h$) are applied to generate component-word relevance matrix ($\boldsymbol{S}^h$) and label-component association matrix ($\boldsymbol{R}^h$) respectively, based on which the label-based word attention ($\boldsymbol{A}^h$) can be generated. Then $\boldsymbol{A}^h$ and $\boldsymbol{I}_{X_i}$ are combined to generate the label-based document embeddings ($\boldsymbol{D}_{X_i}^h$). Besides, local document embedding ($\tilde{\boldsymbol{v}}^{h-1}_{X_i}$) generated from level $h-1$ are applied to filter out unimportant labels to generate the processed label-based document embedding ($\tilde{\boldsymbol{D}}^h_{X_i}$). The input of this Label-based Attention Module are highlighted with dotted boxes which include $\tilde{\boldsymbol{v}}^{h-1}_{X_i}$ as well as $\boldsymbol{I}_{X_i}$ and the output are highlighted with solid boxes which include $\tilde{\boldsymbol{D}}^h_{X_i}$ as well as $\boldsymbol{v}^{h}_{X_i}$.}
                \label{fig:attention module}
            \end{figure}

        This module is designed for performing label-based attention and learning both local and global document features based on different labels at each hierarchical level. However, there are some challenges that need to be addressed:
        \begin{itemize}
            \item Huge number of labels: Due to the large number of labels in some datasets, having multiple attention (i.e., one for each label) is memory intensive. Besides,  some labels are only associated with very few training samples and in these cases, it is difficult to learn high quality attention parameters.
            \item Label dependencies: There could exist dependencies among sibling labels and ignoring such dependencies among labels leads to inefficient learning and information loss.
        \end{itemize}
        From the above challenges, learning independent attention for each individual label is clearly disadvantageous. Hence, we introduce the label-based attention module, where component embeddings function as a factorizer to  bridge the labels and document content. This module mainly consists of two sub-modules: 1) Component-Word Relevance and 2) Label-Component Association. Besides, to effectively leverage information extracted from previous levels, we also proposed 3) Hierarchical Information Integration sub-module. The details of the label-based attention module is as shown in Fig. \ref{fig:attention module}.       
        
        \paragraph{\textbf{Component-Word Relevance}} This module is designed to learn the latent features between words and component at level $h$ to reflect the important information in the input text. 
        Here, components at the level $h$ are represented as a set of vectors: $\mathcal{C}^h=\{\boldsymbol{v}_{c_1}^h,\boldsymbol{v}_{c_2}^h \dots, \boldsymbol{v}_{c_m}^h \}$, which are randomly initialized trainable parameters. The relevance between word $i$ and component $j$ is denoted as $\boldsymbol{S}_{i,j}^h$, which can be calculated through:
            \begin{equation}
            \label{eq:relevance_m}
                \boldsymbol{S}_{i,j}^h = f_w(\boldsymbol{v}^h_{w_i})^T \boldsymbol{v}^h_{c_j}
            \end{equation}
            where $f_w(\cdot)$ is a single-layer MLP and $\boldsymbol{S}^h \in R^{N \times |\mathcal{C}^h|}$ denotes the component-word relevance value matrix. $\boldsymbol{v}^h_{w_i}$ represents the word embedding of word $w_i$ at the level $h$ and more details will be introduced later.
        
        \paragraph{\textbf{Label-Component Association}}
            The objective of this module is to learn the latent relations between each label and the components. Labels at the level $h$ are represented with a set of vectors (label embeddings): $\mathcal{L}^h = \{\boldsymbol{v}_{l_1}^h, \boldsymbol{v}_{l_2}^h,\dots \boldsymbol{v}_{l_q}^h\}$, which are also trainable parameters. The relevance between labels and components $\tilde{\boldsymbol{R}}^h \in R^{|\mathcal{L}^h| \times|\mathcal{C}^h|}$ is calculated as:
            \begin{equation}
            \label{eq:association_m}
                \tilde{\boldsymbol{R}}^h = softmax(\boldsymbol{R}^h)
            \end{equation}
            where $softmax(\cdot)$ is row-based softmax operation. The relevance value between each pair of label $i$ and component $j$ is calculated as:
            \begin{equation}
                 \boldsymbol{R}_{i,j}^h = f_l(\boldsymbol{v}^h_{l_i})^T \boldsymbol{v}^h_{c_j}
            \end{equation}
            where $f_l(\cdot)$ is a single-layer MLP to transform label embeddings to the same domain as component embeddings.
            
            With the relevance matrix $\boldsymbol{S}^h$ and association matrix $\tilde{\boldsymbol{R}}^h$ defined in Eq. \ref{eq:relevance_m} and Eq. \ref{eq:association_m}, we can now compute the label-based word attention matrix $\boldsymbol{A}^h \in R^{|\mathcal{L}^h| \times N}$ through:
            \begin{equation}
                \boldsymbol{A}^h =softmax(\tilde{\boldsymbol{R}}^h{\boldsymbol{S}^h}^T)
            \end{equation}
            
            It is worth noting that the labels are related through their common components. Besides, since component embeddings are trained via all training samples regardless of the labels, it helps to address the challenge of poor quality embeddings due to having too few samples for certain labels. Then, the label-based document embeddings $\boldsymbol{D}_{X_i}^h \in R^{|\mathcal{L}^h| \times d}$ can be generated through a single-layer MLP :
            \begin{equation}
                \boldsymbol{D}_{X_i}^h = \phi (\boldsymbol{A}^h\boldsymbol{I}_{X_i}  \boldsymbol{W}^h + \boldsymbol{b})
            \end{equation}
            where $\boldsymbol{I}_{X_i} \in R^{N \times d}$ denotes the word embedding matrix of the document $X_i$ and the $i$th row of $\boldsymbol{I}_{X_i}$ represents the word embedding $\boldsymbol{v}_{w_i}$ of the $i$th word $w_i$. $\boldsymbol{W}^h \in R^{d \times d}$ and $\boldsymbol{b} \in R^{1 \times d}$ are trainable parameters and $\phi(\cdot)$ denotes an activation function (we choose to use RELU \citep{relu'2012} here). Recall that separate document embeddings will be generated for the global and local classifiers and more details will be introduced in Section \ref{sec:classification}.
            
        \paragraph{\textbf{Hierarchical Information Integration}} 
            Intuitively, hierarchical classifications are carry out from high to low granularity. When we extract label-based document features at the level $h$, we will also refer to the features generated from a higher granularity level (i.e., $h-1$). In the following, we will discuss the two strategies we have employed to integrate the hierarchical information:
            
            1) \textbf{Label-based Document Embedding Masking}: Given the local document embedding $\tilde{\boldsymbol{v}}^{h-1}_{X_i} \in R^{d \times 1}$ from level $h-1$, we can compute $\boldsymbol{m}^h_{X_i}$ the confidence score of each label at the level $h$, where each element of $\boldsymbol{m}^h_{X_i}$ is predicted from $\tilde{\boldsymbol{v}}^{h-1}_{X_i}$ and it can be calculated through:
            
            \begin{equation}
                \boldsymbol{m}^h_{{X_i},j} = sigmoid(f_m(\tilde{\boldsymbol{v}}^{h-1}_{X_i})^T\boldsymbol{v}^{h}_{l_j})
            \end{equation}
            where $f_m(\cdot)$ represents a single-layer MLP. It is worth noting that $\mathbf{m}^1_{X_i} =\mathbf{1} \in R^{|\mathcal{L}^1|\times1}$, since there is no local document embedding from the previous level for the label-based attention module at the level $1$. To avoid over-filtering, LeakyRelu\citep{he2015delving} is chosen as the activation function of $f_m(\cdot)$. 
            With that, the label-based document embeddings can be processed via:
            \begin{equation}
                \tilde{\boldsymbol{D}}^h_{X_i} = mask(\boldsymbol{D}^h_{X_i},\boldsymbol{m}^h_{X_i})
            \end{equation}
            where $mask(\cdot)$ denotes an masking operation that multiplies each row of $\boldsymbol{D}^h_{X_i}\in R^{|\mathcal{L}^h| \times d}$ with the corresponding value in $\boldsymbol{m}^h_{X_i}\in R^{|\mathcal{L}^h| \times 1}$.
            Now, we can generate both the local $\tilde{\boldsymbol{v}}^h_{X_i}$  and global $\boldsymbol{v}^h_{X_i}$ document embeddings at the level $h$:

            \begin{equation}
                 {\tilde{\boldsymbol{v}}^h_{X_i}} = average(\tilde{\boldsymbol{D}}^h_{X_i}{}^T)
            \end{equation}
            \begin{equation}
                 {\boldsymbol{v}^h_{X_i}} = average({\boldsymbol{D}^h_{X_i}}{}^T)
            \end{equation}
            where $average(\cdot)$ denotes the row-based average operation. 
            
            2)\textbf{Word Embedding Enrichment}: When we calculate the component-word relevance matrix $\boldsymbol{S}^h$, the local document embedding from the level $h-1$ will be concatenated with each word embedding of the input document to generate the word embeddings at the level $h$, formally:
            \begin{equation}
                \boldsymbol{v}^h_{w_j} = \tilde{\boldsymbol{v}}^{h-1}_{X_i} \mathbin\Vert \boldsymbol{v}_{w_j} 
            \end{equation}
            
            % So far, we have introduced the procedure of generating label-based hierarchical document embeddings through the label-based attention module. In the following sub-section we will elaborate on the local and global classifications.
            
    \subsection{Classification}
    \label{sec:classification}
    Given the local and global document embeddings, we can simultaneously perform the local classification and global classification.
    \paragraph{\textbf{Local Classification}}
    To preserve the label-based information, we employ the label-based document embeddings $\tilde{\boldsymbol{D}}_{X_i}^h$ instead of the document embedding $\tilde{\boldsymbol{v}}_{X_i}^h$ (information lost due to the averaging operation) to perform the local classification. Formally, at the level $h$, the probability that an input document $X_i$ is associated to the label $l_j$ can be calculated as:
    \begin{equation}
        p_{X_i,j} = sigmoid( \tilde{\boldsymbol{D}}^h_{X_i,j}\boldsymbol{v}_{l_j}^h)
    \end{equation}
    where $\tilde{\boldsymbol{D}}^h_{X_i,j}\in R^{1\times d}$ represents the $j$th row of the label-based document embedding matrix. Then the local loss is measured as follows:
    \begin{equation}
    \begin{aligned}
        \mathcal{O}_{local} = -\sum_{i=1}^{K}\sum_{\substack{h=1, \\ \psi(l_j) =h} }^{H}(z_{l_j}\cdot\log(p_{X_i,j}) \\
        + (1 - z_{l_j})\cdot\log(1 - p_{X_i,j}))
    \end{aligned}
    \end{equation}
    where $z_{l_j} = 1$ iff the document $X_i$ belongs to the label $l_j$ otherwise $z_{l_j} = 0$.
    \paragraph{\textbf{Global Classification}}
    It may be intuitive to perform the classification sequentially, however, in this case the misclassification errors from local classification at each hierarchy level will be propagated to the next level and eventually the global classification. Therefore, to reduce the effect of error-propagation introduced from local classifiers and make full use of the information extracted from other levels, we simultaneously optimize both local and global losses. For the global classification, the global document embeddings extracted from all levels are applied to generate the final global document embedding $\boldsymbol{v}^g_{X_i}$, formally:
    \begin{equation}
        \boldsymbol{v}^g_{X_i} = (\boldsymbol{v}^1_{X_i} \mathbin\Vert \boldsymbol{v}^2_{X_i} \dots \boldsymbol{v}^H_{X_i}) 
    \end{equation}
    The global loss is measured as follows:
    \begin{equation}
    \begin{aligned}
         \mathcal{O}_{global} = -\sum_{i=1}^{K}\sum_{j=1}^M(z_{l_j}\cdot\log(p^{g}_{X_i,j}) \\
        + (1 - z_{l_j})\cdot\log(1 - p^g_{X_i,j}))
    \end{aligned}
    \end{equation}
    where $p^g_{X_i,j}$ is the $j$th element of the global prediction results and $\boldsymbol{p}^g_{X_i} \in R^{M \times 1}$ and $M = \sum_H|\mathcal{L}^h|$ represents the probability that document $X_i$ is associated to the label $l_j$. The global prediction results are calculated through:
    \begin{equation}
        \boldsymbol{p}^g_{X_i} = f_g(\boldsymbol{v}_{X_i}^g)
    \end{equation}
    where $f_g(\cdot)$ is a two-layer MLP.
    
    So, the final loss to be optimized is given as:
    \begin{equation}
        \mathcal{O}_{all} = \mathcal{O}_{local} + \mathcal{O}_{global}
    \end{equation}
    Then the final prediction results can be generated by combining both local and global prediction results which is:
    \begin{equation}
        \boldsymbol{p}^{all}_{X_i}= \alpha\boldsymbol{p}_{X_i} +(1-\alpha)\boldsymbol{p}^{g}_{X_i}
    \end{equation}
    where
    \begin{equation}
        \boldsymbol{p}_{X_i} =  (\boldsymbol{p}^1_{X_i} \mathbin\Vert \boldsymbol{p}^2_{X_i} \dots \boldsymbol{p}^H_{X_i}) 
    \end{equation}
    and $\alpha$ is a predefined parameter to regulate the importance given to the local and global predictions, which we set as $0.5$ here.

\section{Experiments}
    \subsection{Experiment Setup}
    
    \paragraph{\textbf{Datasets}} 
    We evaluate the proposed model on four public HMTC datasets with different properties, WIPO-alpha\footnote{WIPO-alpha is available at \url{ https://www.wipo.int/classifications/ipc/en/ITsupport/Categorization/dataset/index.html}}, BlurbGenreCollection(BGC)\footnote{BGC is available at \url{https://www.inf.uni-hamburg.de/en/inst/ab/lt/resources/data/blurb-genre-collection.html}} \citep{CapsNetwork'2019}, Enron \citep{Klimt04:proc} and Reuters \citep{Lewis04:jrnl}. The statistics of these datasets are presented in Table \ref{table:datasets}.
    
    Note that for BGC, Enron, and Ruters, the lowest granularity labels of each input sample are not necessarily the leaf nodes, which makes them more difficult to be analyzed. While the WIPO-alpha is the only mandatory-leaf \citep{bi_mandatory'2012} dataset,  its large number of labels increases its difficulty.
        \begin{table*}
        \centering
        \begin{tabular}{cclccc}
        \hline Dataset & $|L|$ & Hierarchy & Training & Validation & Testing\\ \hline
        WIPO-alpha & 5,229 & 8, 114, 451, 4,656 & 45,105 & 15,036 & 15,036 \\
        BGC & 146  & 7, 46, 77, 16 & 55,136 & 18,378 & 18,378 \\
        Enron & 56 & 3, 40, 13 & 988 & 330 & 330 \\
        Reuters & 101 & 4, 55, 42 & 3,600 & 1,200 & 1,200\\
        \hline
        \end{tabular}
        \caption{\label{table:datasets} Statistics of experimental datasets. $|L|$ denotes the total number of labels in the label hierarchy. \textit{Hierarchy} denotes the number of labels at each of the hierarchical level, e.g. the dataset WIPO-alpha includes four levels of labels, and the number of labels at each level is 8, 114, 451, and 4,363, respectively.}
        \end{table*}
    
    \paragraph{\textbf{Evaluation Metrics}}
    For each input sample, the model outputs a number of scores, where each score indicates the probability that the input sample belongs to the corresponding category. To employ evaluation measures such as precision, recall, or f1-score, it is necessary to select a threshold for qualitative classification based on the scores. However, the best threshold value may differ depending on the dataset and it is not the focus of this work. Therefore, we employ the area under the average precision-recall curve ($AU(\overline{PRC}$) measure, which is also widely acceptable in HMTC domain \citep{lmlp'2016}, to evaluate the performance of the algorithms. 
    
    \subsection{Baseline Methods}
        To provide comprehensive evaluation, we compare our proposed method with four state-of-the-art neural network-based HMTC algorithms, as well as three variants of LA-HCN. 
        
        \begin{itemize}
            \item \textbf{HMCN-F}, \textbf{HMCN-R} \citep{hmcn'2018}: The feedforward (F) and recurrent (R) version of HMCN, which is the first neural network-based HMC method that combines both local and global information to perform hierarchical classification. During the training procedure, label hierarchy information is preserved by penalizing hierarchical violations.
            \item \textbf{Cap.Network} \citep{CapsNetwork'2019}: It is the first work that introduces capsules into HMC tasks. By associating each label in the hierarchy with a capsule and applying routing mechanism, Cap.Network indicates the effectiveness of capsules in feature identification and combination. The label hierarchy information is directly applied to perform post-processing (i.e., label correction) during the prediction stage.
            \item \textbf{HARNN} \citep{HARNN'2019}: It is the most related algorithm to LA-HCN which also optimizes both local and global loss simultaneously. Besides, attention mechanism is applied at each hierarchical level to extract level-based document features for better local classification performance. At the prediction stage, local and global prediction scores are combined to generate the final results.
        \end{itemize}
        
        The selected baselines are all state-of-the-art neural network-based HMTC algorithms which cover both flat approach (Cap.Network) and hybrid-approaches (HARNN, HMCN). 
        
        \subsection{Ablation Study}
        To demonstrate the effectiveness of our proposed label-based attention mechanism and the combination of local and global approaches, three variants of LA-HCN are also tested. Here, LA-HCN$_{NC}$ (NC stands for no component) represents the variant of LA-HCN with the component embedding replaced by simple dot-product. LA-HCN$_G$ and LA-HCN$_L$ represent the variants of LA-HCN which only optimize global and local loss, respectively. For fair comparison, the input features of HMCN are replaced with the output of Bi-LSTM or pre-trained word embeddings, which leads to slightly better performance of HMCN than what is reported in their paper.
        
    \subsection{Implementation Details}
    % Training, testing and validation split is already provided for BGC, Enron and Reuters so we just follow the original settings. As for WIPO-alpha, we randomly select $20\%$ of samples as testing and validation sets respectively.
    
    Given that the raw text of WIPO-alpha and BGC are available, we apply Bi-LSTM --- a variant of LSTM \citep{lstm'1997} as the base text encoder of these two datasets. However, since only processed word statistic information is provided for Enron and Reuters, so we simply use pre-trained word embeddings as the input text features. 
    % More details can be find in the Appendix \ref{appends:Implementation Details}.

    \subsection{Experimental Results}
    \paragraph{\textbf{Classification Results}}
    We compare the performance of LA-HCN and its variants with other baseline methods and the results are shown in Table \ref{table:overall performance}. 
    \begin{table}
        \centering
        \begin{tabular}{c|cccc}
        \hline Algorithms & WIPO-alpha & BGC & Enron & Reuters \\ \hline
        Cap.Network & OOM & 0.7827 & 0.7552 & 0.7119 \\
        HMCN-F & 0.5487 & 0.7983 & 0.7795 & 0.7505 \\
        HMCN-R & 0.5512 & 0.7968 & 0.7820 & 0.7339 \\
        HARNN & 0.5817 & 0.8214 & 0.7862 & 0.7528\\ \hline
         LA-HCN$_{NC}$ & 0.5564  & 0.8148 & 0.7859 & 0.7482 \\
         LA-HCN$_L$ & 0.5736 & 0.8231  & 0.7870 & 0.7619 \\
         LA-HCN$_G$ & 0.5930  & 0.8224 & 0.7994 & \textbf{0.8101} \\
         LA-HCN &\textbf{0.5965} & \textbf{0.8277}  & \textbf{0.8015} & 0.8006 \\
        \hline
        \end{tabular}
        \caption{\label{table:overall performance} $AU(\overline{PRC}$) comparison on the four datasets. OOM denotes out of memory.}
     \end{table}
     
    %  \begin{table}
    %     \centering
    %     \begin{tabular}{c|cccc}
    %     \hline Algorithms & \makecell[c]{WIPO-\\alpha} & BGC & Enron & Reuters \\ \hline
    %     HMCN-F & 0.517 & 0.805 & 0.721 & 0.675 \\
    %     HMCN-R & 0.528 & 0.804 & 0.739 & 0.674 \\
    %     Cap.Network & OOM & 0.7613 & 0.713 & 0.6562 \\
    %     HiLAP &  &  &  &  \\
    %     HARNN & 0.573 & 0.822 & 0.745 & 0.667\\
    %     \textbf{-Local} &  & 0.815  & 0.726 & 0.645 \\
    %     \textbf{-Global} &  &  0.825 & 0.676 & 0.725 \\
    %     \textbf{-} &\textbf{ 0.590} & \textbf{0.830}  & \textbf{0.749} & \textbf{0.740} \\
    %     \hline
    %     \end{tabular}
    %     \caption{\label{table:overall performance} Performance comparison on the four datasets. (Evaluation Metric:$AU(\overline{PRC}$)  }
    %  \end{table}
     
     LA-HCN outperforms the baselines on all the datasets and more significantly on Reuters, indicating the effectiveness of LA-HCN in HMTC problem. More specifically, the hybrid-approaches (i.e., HMCN, HARNN, LA-HCN) outperform simple flat classification algorithms such as Cap.Network on all datasets, which shows the advantages of having a combination of  both local and global classifications. The performance of HARNN and LA-HCN is better than HMCN on all four datasets, which indicates that hierarchically extracting document features based on label levels is helpful to improve the hierarchical classification performance. By comparing the LA-HCN variants, we observe that LA-HCN without the component embeddings (LA-HCN$_{NC}$) performed the worst among its variants. This shows the importance of the component embeddings in addressing the huge number of labels and label dependencies. We also observe that the performance of global classification is typically better (three out of four cases) than that of local classification and combining local and global typically outperforms individual (i.e., local only or global only). Separately generating document embeddings for local and global classification is one of the main reasons why LA-HCN performs better than other algorithms where the combined local and global classifications are also applied. The separation of global document embeddings in LA-HCN can effectively reduce the error-propagation introduced by local classification. Moreover, the performance of LA-HCN can  be further improved with local label-based document embeddings which provide additional hierarchical information. 
    %  Besides, usually when the label structure is more complicated, the improvement becomes more obvious (WIPO-alpha).

     \paragraph{\textbf{Learned Label-based Attention}}
     From the above experiments, we can see that one of the most important characteristic of LA-HCN is its ability to capture important information of document based on different labels. In this experiment, we visualize the learned attention based on labels from different levels to verify its effectiveness. The BGC dataset is used to illustrate the experimental results. We also show the learned attentions from HARNN and HAN for the comparisons. 
     
    \begin{table}
        \centering
        \begin{tabular}{c|ccc}
        \hline Dataset & LA-HCN & HARNN & HAN \\ \hline
        BGC & 0.8277 & 0.8214 & 0.8153 \\
        \end{tabular}
        \caption{\label{table:visualization}The $AU(\overline{PRC})$ of LA-HCN, HARNN and HAN on BGC dataset.}
     \end{table}
     
     \begin{figure*}
        \centering
        \includegraphics[width=4.75in]{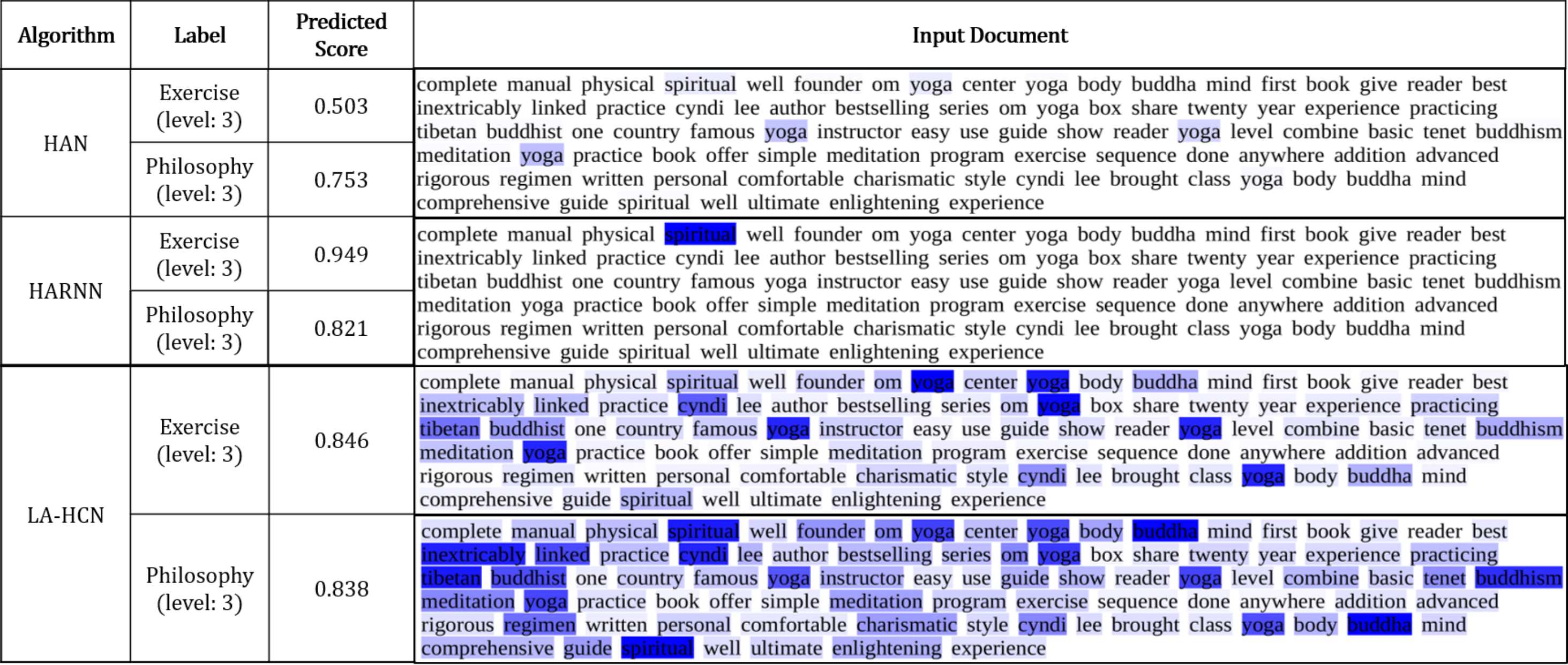}
        \caption{Visualization of the learned attention (words) at the level $h=3$ from HAN, HARNN and LA-HCN. The corresponding labels as well as the prediction probabilities are also presented.}  
        \label{fig:attention}
    \end{figure*}
    
    \begin{figure*}
        \centering
        \includegraphics[width=4.75in]{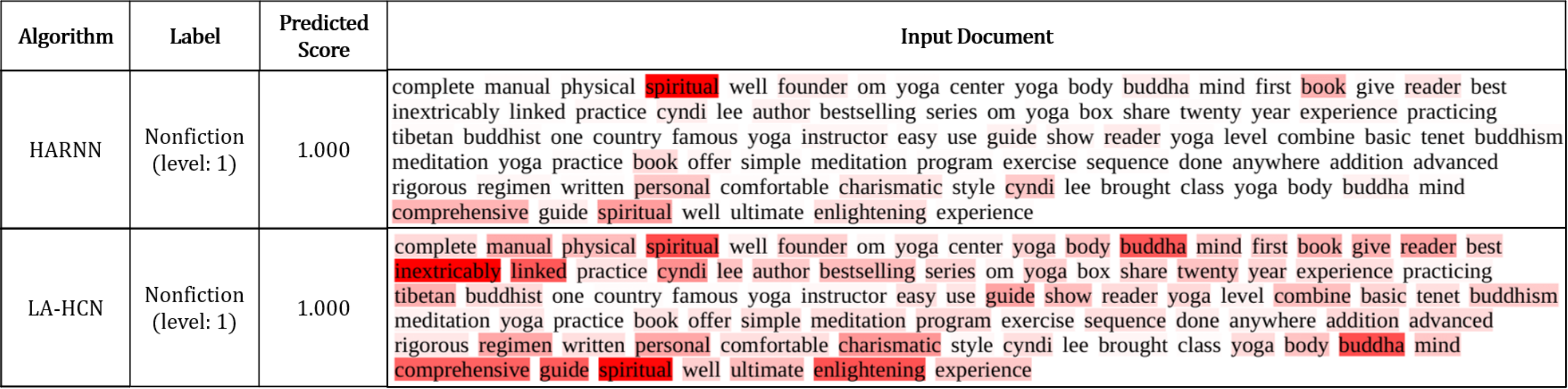}
        \caption{Visualization of the learned attention(words) at the level $h=1$ from HARNN and LA-HCN. The corresponding labels as well as the prediction probabilities are also presented.}  
        \label{fig:attention2}
    \end{figure*}
    
     The classification performance of the three algorithms on BGC is shown in Table \ref{table:visualization} and the learned attentions at two different hierarchical levels are shown in Fig. \ref{fig:attention} and Fig. \ref{fig:attention2}, respectively. 
     
     In Fig. \ref{fig:attention}, we show the attention learned from the level $h=3$ where the text is associated with the label \textit{Exercise} and \textit{Philosophy}. Clearly, LA-HCN pays more attention to the word ``\textit{yoga}'' when the corresponding label is \textit{Exercise} while more attention is paid to ``\textit{spiritual}'', ``\textit{buddha}'' and ``\textit{tibetan}'' when the label is \textit{Philosophy}. However, HARNN and HAN can only generate the same attention value for all labels which is difficult to understand in some case. 
     
     In Fig. \ref{fig:attention2}, we show the attention learned based on the label \textit{Nonfiction} from the level $h=1$. Combined with the attention learned from the level $h=3$ in Fig. \ref{fig:attention}, we can find that, in LA-HCN, the word "\textit{yoga}" is not important at the level $h=1$, where it is associated with the label \textit{Nonfiction}, while it can still be captured at the level $h=3$ when it is associated with the label \textit{Exercise}. However, HARNN can only select important words at the level $h=3$ based on the attention learned from its previous level so that the word "\textit{yoga}" will be lost at the level $h=3$ since "\textit{yoga}" is not important for the classification at the level $h=1$. This shows that the sharing of hierarchical information in LA-HCN is more effective.

\section{Conclusion}
We proposed LA-HCN, a novel algorithm for HMTC, where label-based attention are learned for the text at different hierarchical levels. Furthermore, both local and global text embeddings are generated for local and global classification respectively. At different levels, meaningful attention can be learned based on different labels. Comprehensive experiments demonstrate the effectiveness of LA-HCN, which outperforms other neural network-based state-of-the-art HMTC algorithms across four benchmark datasets of different properties. Moreover, the visualization of the learned label-based attentions reveals its interpretability. However, the practical meaning of learned components is not well explored so far and may be considered in our future work, where the explicit label structure may also be taken into consideration in the design of the components. 

\section*{Acknowledgement}
This research/project is supported by the National Research Foundation, Singapore under its AI Singapore Programme (AISG Award No: AISG-100E-2019-031). Any opinions, findings and conclusions or recommendations expressed in this material are those of the author(s) and do not reflect the views of National Research Foundation, Singapore.

\bibliography{HMC_report}

\end{document}